\documentclass[lettersize,journal]{IEEEtran}
\usepackage{amsmath,amsfonts}
\usepackage{algorithmic}
\usepackage{algorithm}
\usepackage{array}
\usepackage[caption=false,font=normalsize,labelfont=sf,textfont=sf]{subfig}
\usepackage{textcomp}
\usepackage{stfloats}
\usepackage{url}
\usepackage{verbatim}
\usepackage{graphicx}
\usepackage{cite}
\usepackage{booktabs}
\usepackage{multirow}
\begin{document}

\title{M3SunAgent: Monocular 3D Spatial Understanding Agent for Metric Depth Estimation and 3D Visual Grounding}

\author{
  Jinsong~Zhang,
  Kejun~Wu,~\IEEEmembership{Senior Member,~IEEE},
  Ming~Zhu,~\IEEEmembership{}
  Renjie~Qiao,~\IEEEmembership{}
  Chengtao~Cai,~\IEEEmembership{Senior Member,~IEEE},
  and Zhengguo~Li,~\IEEEmembership{Fellow,~IEEE}
  \thanks{Jinsong~Zhang, Kejun~Wu, and Ming~Zhu are with the School of Electronic Information and Communications, Huazhong University of Science and Technology, Wuhan 430074, China.}
  \thanks{Renjie~Qiao and Chengtao~Cai are with College of Intelligent Systems Science and Engineering, Harbin Engineering University, Harbin 150001, China.}  
  \thanks{Zhengguo~Li is with 
  Institute of Advanced Intelligence and Computation, Agency for Science, Technology and Research (A*STAR), Singapore.}
  \thanks{This work was supported by the National Natural Science Foundation of China under Grant 62501246.}
  }

\maketitle

\begin{abstract}
Monocular metric depth estimation and 3D visual grounding represent the two complementary cornerstones of monocular 3D spatial understanding (M3Sun), from which the fundamental 3D spatial information required by M3Sun can be acquired. 
However, these complementary tasks are generally conducted by separate frameworks, which pose challenges of inflexible and unaligned spatial information access for embodied intelligence systems.
In this paper, we propose a unified agent for monocular 3D spatial understanding (\textbf{M3SunAgent}) that leverages a large language model (LLM) as a task planner for spatial visual programming, which flexibly generate structured programs and coordinate tools.
For instance-level metric depth estimation task, M3SunAgent invokes an object detector tool to locate the target, estimates depth at selected points with a depth estimation tool, and aggregates these predictions into an instance-level depth estimate. We also construct the M3Sun Instance (M3SI) dataset, a benchmark with 2,910 samples for evaluation.
For monocular 3D visual grounding task, M3SunAgent uses a vision-language model (VLM) tool to locate the target and output basic spatial attributes, then combines back-projection tool
with a dimension-lifting tool to predict its 3D bounding box. 
Experimental results demonstrate the superior performance of M3SunAgent. Specifically, in evaluations of instance-level monocular metric depth estimation, M3SunAgent achieves the best performance among all compared models, $52.61\%$ predicted instances are distributed below depth error 0.25 ($\delta_{<0.25}$).
In evaluations of monocular 3D visual grounding, M3SunAgent demonstrates overall competitive performance than vision and VLM models, reaching a 3D mean intersection over union (mIoU) of $41.73\%$ and exceeding the state-of-the-art MonoVLM model by $3.62\%$.
\end{abstract}


\section{Introduction}
\label{sec1}
\IEEEPARstart{M}{onocular} 3D spatial understanding (M3Sun) aims to understand the spatial relationships in an image, involving recognizing objects, and estimating their distances, 3D locations, physical extents, and orientations. The precise spatial properties of individual objects provided by M3Sun plays a vital role in embodied intelligence, teleoperation, and autonomous driving.
For example, when a robot is instructed to ``pick up the pen on the table'',  it needs to locate the referred pen in the input image and estimate the distance to guide its motion planning~\cite{spatialvlm}. In autonomous driving, an instruction such as ``Avoid the car ahead on the right'' requires the system to identify the referred vehicle and estimate its 3D position, dimensions, and heading. Such scenarios require systems to locate objects described in natural language and infer their spatial properties, motivating continued advances in M3Sun.

Instance-level monocular metric depth estimation and monocular 3D visual grounding represent the two cornerstones of M3Sun. These tasks offer complementary 3D spatial information, e.g., instance-level depth and 3D bounding box. 
The former estimates the metric depth of an instance in physical units, while the latter identifies the referred object and recovers its 3D structure, including its position, dimensions, and orientation~\cite{mono3dvg}. Both types of spatial information are important for understanding objects from a single image. Metric depth places object distance on a physical scale, enabling quantitative distance judgments. Accurate metric depth estimation can also provide additional cues for spatial reasoning. 3D structure characterizes the object's location and spatial extent. When accurate 3D structures are provided as auxiliary cues, they can improve reasoning about object arrangement and spatial relationships~\cite{wang2026monosr}. Both tasks require systems to ground language descriptions in images and infer quantitative spatial properties by combining visual evidence with semantic cues. The common demand to identify objects and produce quantitative spatial estimates motivates studying these two tasks together.

Recent research in M3Sun has explored various approaches to improve the accuracy of estimated spatial properties. Training on diverse scenes and modeling camera parameters have improved metric distance estimation~\cite{hu2024metric3d}. Beyond scale and camera modeling, planar geometry and scene semantics provide additional cues for monocular depth estimation~\cite{liu2025plane2depth,zhu2026scaledepth}. These developments provide a basis for estimating the distance to a specified object after its location has been identified in the image. Improving estimates of the target's 3D location and extent requires better alignment between language descriptions and image geometry. Spatial descriptions can provide cues for predicting the target's 3D structure~\cite{li2026Mono3DVG-EnSD}. Target localization is also fundamental to this task. Accurately identifying the referred object in the image helps focus geometric estimation on that object and reduce the influence of irrelevant scene elements~\cite{monovlm}. Together, these advances establish a strong basis for recovering object-level spatial information from a single RGB image.

Estimating the distance to a referred object requires locating the target and extracting depth measurements from the corresponding instance. However, many monocular depth estimators predict depth across the entire image~\cite{shao2025monodiffusion,hu2024metric3d}. Such scene-level predictions do not identify the referred object or directly provide its instance-level depth. In monocular 3D visual grounding, specialized architectures have achieved strong performance~\cite{li2026Mono3DVG-EnSD}. However, models designed for this specific task may lack the flexibility to handle other spatial queries. Advances in large language models (LLMs) and vision-language models (VLMs) have enabled their application to a wider range of visual tasks~\cite{wu2026corrupted}. These developments have motivated attempts to improve the flexibility of visual models for spatial understanding. With spatial training, VLMs can reason about object positions, orientations, and 3D relationships~\cite{ma2025spatialllm}. However, achieving such performance may require substantial 3D training data, computational resources, and multi-stage training procedures~\cite{song2025robospatial}. Moreover, VLMs may still face challenges in producing accurate metric distance estimates due to the difficulty of grounding visual observations into physical scales~\cite{bhat2026consistent}.

Meanwhile, advances in LLM-based agents provide a potential solution to these limitations. In an agent architecture, the LLM serves as a central cognitive planner that performs reasoning and coordinates tool execution. Specialized models and computational tools execute specific operations and pass intermediate results between steps~\cite{chen2026spacetools}. The agent then integrates their outputs to generate the final response. This division of labor allows the planning LLM to be applied without additional training, reducing the related cost. It also allows the agent to coordinate different expert models within a shared framework and support multiple tasks. Additional models and tools can be incorporated to extend the range of spatial queries that the agent can handle~\cite{shen2023hugginggpt}. Furthermore, quantitative outputs from specialized components can provide the LLM with more reliable evidence for generating the final response. These capabilities have contributed to the growing use of agent-based methods for 3D spatial understanding~\cite{VISPROG}. For these two tasks, a unified agent should retain intermediate measurements of the queried object and combine them to produce the required spatial output.

To address the need for a unified agent in M3Sun, we propose \textbf{M3SunAgent}, a monocular spatial understanding agent that converts a natural language query into an executable program. The LLM generates programs by identifying the requested task and arranging the required operations in the proper order. A program interpreter then executes these operations through registered visual tools and geometric computations while transferring intermediate results between operations. For instance-level metric depth estimation, the agent localizes the referred object and estimates its depth at selected image points. It aggregates these predictions into an instance-level metric distance estimate. For monocular 3D visual grounding, a VLM predicts the target's basic spatial attributes. Back-projection and a specialized 2D-to-3D lifting model use these predictions to recover the 3D geometry required to construct its 3D bounding box. Both pipelines share the same planning and execution framework. Our main contributions are summarized as follows:

\begin{itemize}
    \item We propose \textbf{M3SunAgent}, a monocular spatial understanding agent that flexibly supports instance-level metric depth estimation and 3D visual grounding. M3SunAgent leverages spatial visual programming to understand diverse natural-language queries by a LLM task planner, converting complex queries into executable programs. These programs coordinate distinct models and operations, allowing them to collaboratively execute for spatial understanding.
 
    \item For metric depth estimation, we use focal-length normalization to reduce the effects of intrinsics, together with multi-point sampling and outlier filtering to improve robustness. For monocular 3D visual grounding, we use a VLM for target localization and basic spatial attribute prediction, explicit back-projection for camera adaptation, and a specialized lifting model for accurate dimensions and yaw estimation.

    \item We construct the \textbf{M3Sun Instance (M3SI)} dataset, a benchmark with 2,910 samples for instance-level metric depth estimation. We evaluate M3SunAgent on M3SI for this task and on Mono3DRefer for monocular 3D visual grounding. Across these evaluations, the agent outperforms the spatial VLM baselines and achieves competitive performance to task-specific vision methods.
\end{itemize}

\section{Related Work}
\label{sec2}

\subsection{Monocular metric depth estimation} 

Monocular metric depth estimation aims to recover depth values in physical units from a single RGB image. Its generalization is influenced by camera parameters, scene scale, and imaging conditions. The focus of this task has gradually shifted from improving benchmark performance toward real-world scale recovery, cross-domain generalization, and adaptation to diverse camera models~\cite{shao2025iebins}. Methods that consider focal length and depth scale seek to improve absolute depth prediction in unseen scenes~\cite{wei2024fsdepth}. Some methods estimate camera parameters directly from the input image to improve high-resolution depth prediction~\cite{bochkovskiy2025depthpro}. Self-supervised methods have also addressed challenging conditions, including day--night illumination changes~\cite{hou2024allday}. These advances improve depth prediction across diverse scenes and imaging conditions. However, these depth maps primarily describe scene-level geometry. They do not directly provide the distance to an object identified by language.

Research on monocular 3D detection has also explored object-level depth estimation. Some methods separate objects from background regions and estimate object depth to assist monocular 3D object detection, while others exploit object-related depth information to improve 3D localization~\cite{wang2020Task-aware,peng2022did}. These approaches support 3D localization but are designed for detection rather than instance-level metric depth estimation. At the point level, DepthLM uses visual prompts to estimate metric depth at specified image locations~\cite{depthlm}. A single point may nevertheless be insufficient for an extended or partially occluded object, and a target region may contain background surfaces. Thus, obtaining a stable distance for a specific instance requires both reliable depth measurements from the target and strategies to reduce the influence of unreliable samples.

\subsection{Monocular 3D visual grounding} 
Recent studies have explored aligning language with visual content represented in different forms~\cite{liang2026cibic}. Within this context, visual grounding aims to associate a referring expression with a target object in an image or a 3D scene. In 2D images, visual grounding methods use visual and linguistic information to locate the referred object~\cite{wang2025robustvg}. Grounding in 3D scenes further incorporates geometric information to identify the referred object. ScanRefer matches language descriptions to objects in RGB-D scans~\cite{chen2020scanrefer}. Pseudo-EV addresses ambiguity in viewpoint-dependent spatial descriptions~\cite{geng2025pseudoev}. However, these 3D approaches use observations that are unavailable when only a single RGB image is available.

Monocular 3D visual grounding aims to infer the referred object's 3D location and extent from a single image and language descriptions. Mono3DVG introduced this setting using descriptions that include both appearance and geometric cues~\cite{mono3dvg}. Subsequent methods use cross-modal interaction to incorporate geometric features~\cite{Mono3DVG-TGE} and dimension-decoupled text encoding to represent spatial expressions~\cite{li2026Mono3DVG-EnSD}. More recently, VLM-based approaches have also emerged. MonoVLM adapts a VLM for this task~\cite{monovlm}. GR3D integrates 2D and 3D grounding within a spatial VLM~\cite{cheng2026gr3d}. Although these methods achieve strong performance, they are generally designed for specific tasks. As a result, they may lack the flexibility to handle other types of spatial queries.

\subsection{LLM Agents for Spatial Understanding} 
LLMs can use contextual information to support reasoning~\cite{gao2024contextual}. Agent-based methods further use LLMs to plan operations and coordinate specialized models to complete different tasks. HuggingGPT uses an LLM to plan subtasks and select expert models. It then executes the subtasks with those models and integrates their outputs~\cite{shen2023hugginggpt}. Another line of work represents the required operations as executable programs. VisProg generates such programs from language instructions and composes visual operations to obtain a result~\cite{VISPROG}. For 3D spatial questions, VADAR uses a dynamic API that can be extended with functions for new queries~\cite{VADAR}. SpaceTools further investigates spatial tool coordination by training a VLM through interactive exploration and feedback~\cite{chen2026spacetools}. These studies demonstrate the potential of LLM agents for modular perception and embodied reasoning. However, their application to monocular metric depth estimation and monocular 3D visual grounding remains underexplored. M3SunAgent addresses this gap by using a unified agent framework to coordinate specialized visual models and geometric tools for both tasks. It can be adopted to integrate conventional physics-driven control methods and emerging data-driven vision language action to address long horizon tasks from the hybrid system point of  view ~\cite{li2005switched}.

\section{Method}
\label{sec3}

\subsection{Overview of M3SunAgent}

\begin{figure*}[!htbp]
    \centering
    \includegraphics[width=1\textwidth]{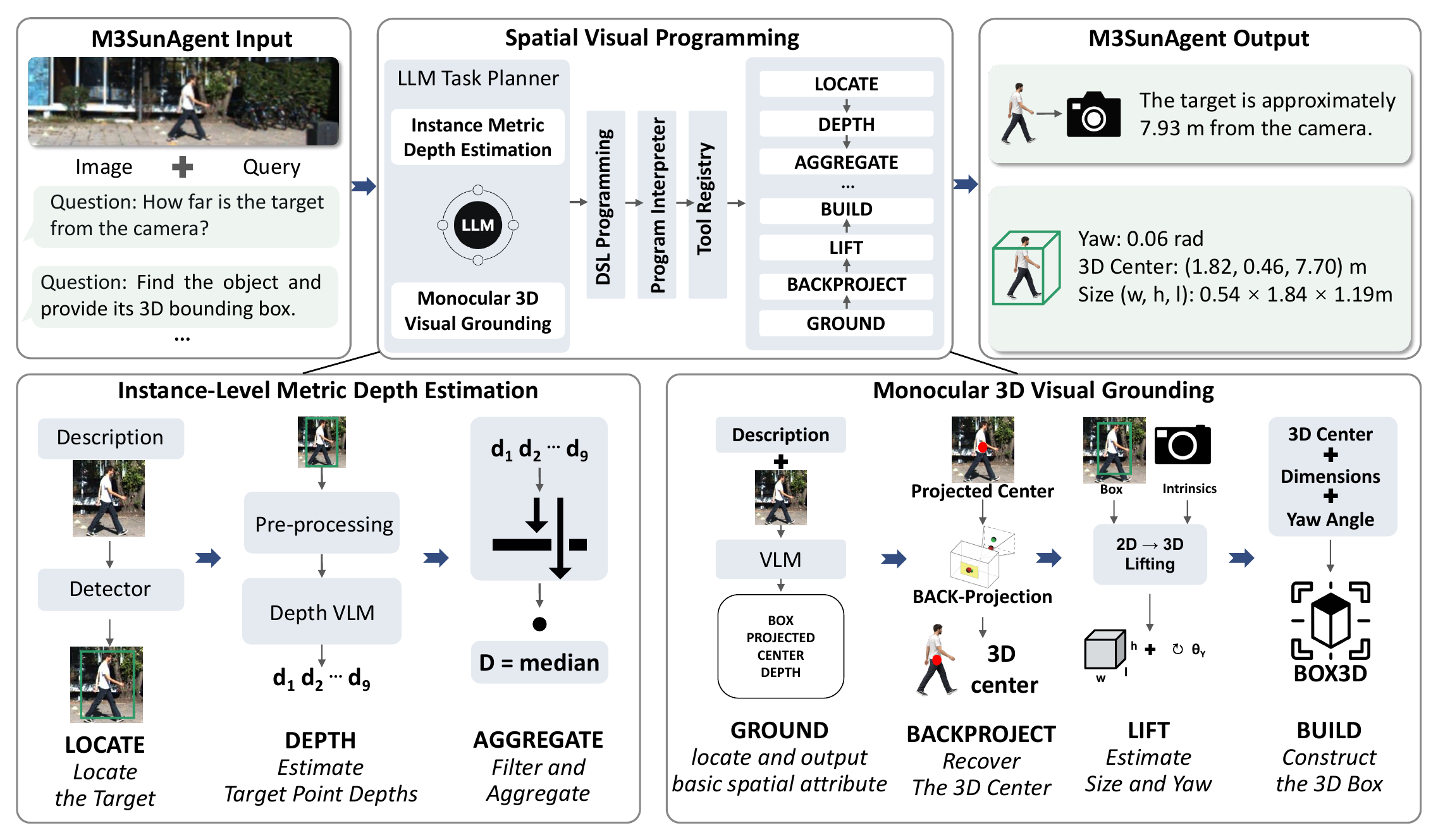}
    \caption{
    Overview of M3SunAgent. Given an RGB image and a natural-language query, the LLM selects the task and generates a corresponding DSL program, which is executed by the program interpreter using perception models and geometric operations. For instance-level metric depth estimation, point-wise depth predictions within the referred object are aggregated into an instance-level metric distance. For monocular 3D visual grounding, a VLM locates the target and predicts basic spatial attribute, which is combined with back-projection and a 2D-to-3D lifting model to construct the 3D bounding box. The result is then used to generate the final response.}
    \label{fig:dga_framework}
\end{figure*}

M3SunAgent takes a single RGB image and a natural-language query as inputs. As shown in Fig.~\ref{fig:dga_framework}, a LLM serves as the task planner, determining whether the query corresponds to instance-level metric depth estimation or monocular 3D visual grounding. Through spatial visual programming, the LLM then generates the corresponding domain-specific language (DSL) program~\cite{VADAR}. The program interpreter then executes the specified operations by invoking registered visual tools and geometric computations from the Tool Registry. The resulting spatial estimate is passed to the LLM, which combines it with the original query to formulate the final answer. This shared planning and execution framework supports both M3Sun tasks.

The two tasks produce different forms of spatial information. A distance query requires a metric estimate associated with the referred instance. A 3D grounding query requires a 3D bounding box describing the instance's position, dimensions, and orientation. In both cases, the DSL program passes target-specific intermediate results between operations, ensuring that subsequent operations remain associated with the same referred object. This keeps subsequent estimates associated with the same object. The following subsections describe the program execution mechanism and the operators used for each task.

\subsection{Spatial Visual Programming}

\label{sec3sub1}

LLMs and VLMs provide strong instruction understanding and semantic reasoning, but reliable metric depth estimation and monocular 3D visual grounding remain challenging. Specialized vision models provide explicit localization and geometric estimates, but are typically designed for predefined perception tasks. To combine these complementary capabilities, M3SunAgent adopts a spatial visual programming paradigm \cite{VISPROG}. The workflows of the two tasks are shown in Algorithms~\ref{alg:prompted_metric_depth} and~\ref{alg:monocular_3d_box}. Given a natural language query, the LLM generates a DSL program composed of task-specific operators. The program interpreter executes these operators by invoking visual tools and geometric computations registered in the Tool Registry for target localization, depth estimation, aggregation, 3D center recovery, geometric attribute estimation, and 3D box construction.

\begin{algorithm}[!htbp]
    \caption{Execution for Instance-Level Metric Depth Estimation}
    \label{alg:prompted_metric_depth}

    \footnotesize

    \noindent\textbf{Notation:}
    $\mathcal{E}$ denotes the in-context DSL examples.
    $P$ denotes the generated DSL program.
    $\mathcal{T}$ denotes the tool registry.
    $r$ denotes the LLM invocation index.
    \par\smallskip

    \begin{algorithmic}[1]
        \FOR{$r\in\{1,2\}$}

            \IF{$r=1$}
                \STATE $\mathrm{LLM}(\textit{QUERY},\mathcal{E})
                \rightarrow
                (\textit{TASK},\textit{DESCRIPTION},P)$

                \IF{$\textit{TASK}=\textit{INSTANCE\_DEPTH}$}
                    \STATE \textit{Generated DSL program $P$:}
                    \STATE \textit{TARGET = LOCATE(IMAGE, DESCRIPTION)}
                    \STATE \textit{DEPTHS = DEPTH(IMAGE, TARGET, INTRINSICS)}
                    \STATE \textit{DISTANCE = AGGREGATE(DEPTHS)}

                    \STATE $\textit{ToolRegistry}
                    \rightarrow \mathcal{T}$

                    \STATE $\textit{ProgramInterpreter}(P,\mathcal{T})
                    \rightarrow \textit{DISTANCE}$
                \ENDIF

            \ELSIF{$r=2 \land
            \textit{TASK}=\textit{INSTANCE\_DEPTH}$}

                \STATE $\mathrm{LLM}
                (\textit{QUERY},\textit{DISTANCE})
                \rightarrow \textit{ANSWER}$

                \RETURN \textit{ANSWER}
            \ENDIF

        \ENDFOR
    \end{algorithmic}
\end{algorithm}

\begin{algorithm}[!htbp]
    \caption{Execution for Monocular 3D Visual Grounding}
    \label{alg:monocular_3d_box}

    \footnotesize

    \noindent\textbf{Notation:}
    $\mathcal{E}$ denotes the in-context DSL examples.
    $P$ denotes the generated DSL program.
    $\mathcal{T}$ denotes the tool registry.
    $r$ denotes the LLM invocation index.
    \par\smallskip

    \begin{algorithmic}[1]
        \FOR{$r\in\{1,2\}$}

            \IF{$r=1$}
                \STATE $\mathrm{LLM}(\textit{QUERY},\mathcal{E})
                \rightarrow
                (\textit{TASK},\textit{DESCRIPTION},P)$

                \IF{$\textit{TASK}=\textit{GROUNDING\_3D}$}
                    \STATE \textit{Generated DSL program $P$:}
                    \STATE \textit{B2D, C2D, DZ = GROUND(IMAGE, DESCRIPTION)}
                    \STATE \textit{C3D = BACKPROJECT(C2D, DZ, INTRINSICS)}
                    \STATE \textit{SIZE, YAW = LIFT(IMAGE, B2D, INTRINSICS)}
                    \STATE \textit{BOX3D = BUILD(C3D, SIZE, YAW)}

                    \STATE $\textit{ToolRegistry}
                    \rightarrow \mathcal{T}$

                    \STATE $\textit{ProgramInterpreter}(P,\mathcal{T})
                    \rightarrow \textit{BOX3D}$
                \ENDIF

            \ELSIF{$r=2 \land
            \textit{TASK}=\textit{GROUNDING\_3D}$}

                \STATE $\mathrm{LLM}
                (\textit{QUERY},\textit{BOX3D})
                \rightarrow \textit{ANSWER}$

                \RETURN \textit{ANSWER}
            \ENDIF

        \ENDFOR
    \end{algorithmic}
\end{algorithm}

M3SunAgent uses in-context examples to guide DSL program generation. We construct several example programs based on the predefined task workflows and provide them to the LLM together with the current query. Rather than directly predicting the final spatial result, the LLM extracts the target description and generates a DSL program based on these examples. Each DSL statement assigns the result of an operator call to an output variable. For example, target localization is expressed as \textit{TARGET = LOCATE(IMAGE, DESCRIPTION)}, where \textit{LOCATE} is the DSL operator, \textit{IMAGE} and \textit{DESCRIPTION} are its input variables, and \textit{TARGET} is its output variable. The output variables can be referenced by subsequent statements, thereby defining the data dependencies within the program. The LLM therefore serves as the task planner rather than directly predicting the requested spatial result.

After generation, the DSL program is parsed and executed by the Program Interpreter~\cite{VISPROG}. The interpreter processes the statements in execution order and retrieves the tool corresponding to each operator from the Tool Registry. A registered tool may be either a learned model or a deterministic computations. After each tool is invoked, its result is stored in the corresponding output variable and passed to subsequent operations when required. For example, the target localized by \textit{LOCATE} can be passed to \textit{DEPTH} to estimate depth values at selected target points. This execution mechanism allows M3SunAgent to connect multiple specialized tools through the intermediate variables defined in the DSL program.

The agent calls the LLM twice for each query. In the first invocation, the LLM uses the query and in-context examples to identify the requested task and generate a corresponding DSL program. The interpreter then executes the program by invoking registered visual tools and geometric operations. Together, these operations produce either the referred object's metric distance or its 3D bounding box. In the second invocation, the LLM combines the computed result with the original query to formulate the final answer. The LLM thus handles task planning and response synthesis, while specialized tools perform spatial estimation. Defining DSL operators separately from their implementations also allows tools to be added or replaced.

\begin{figure*}[!t]
    \centering
    \includegraphics[width=0.95\textwidth]{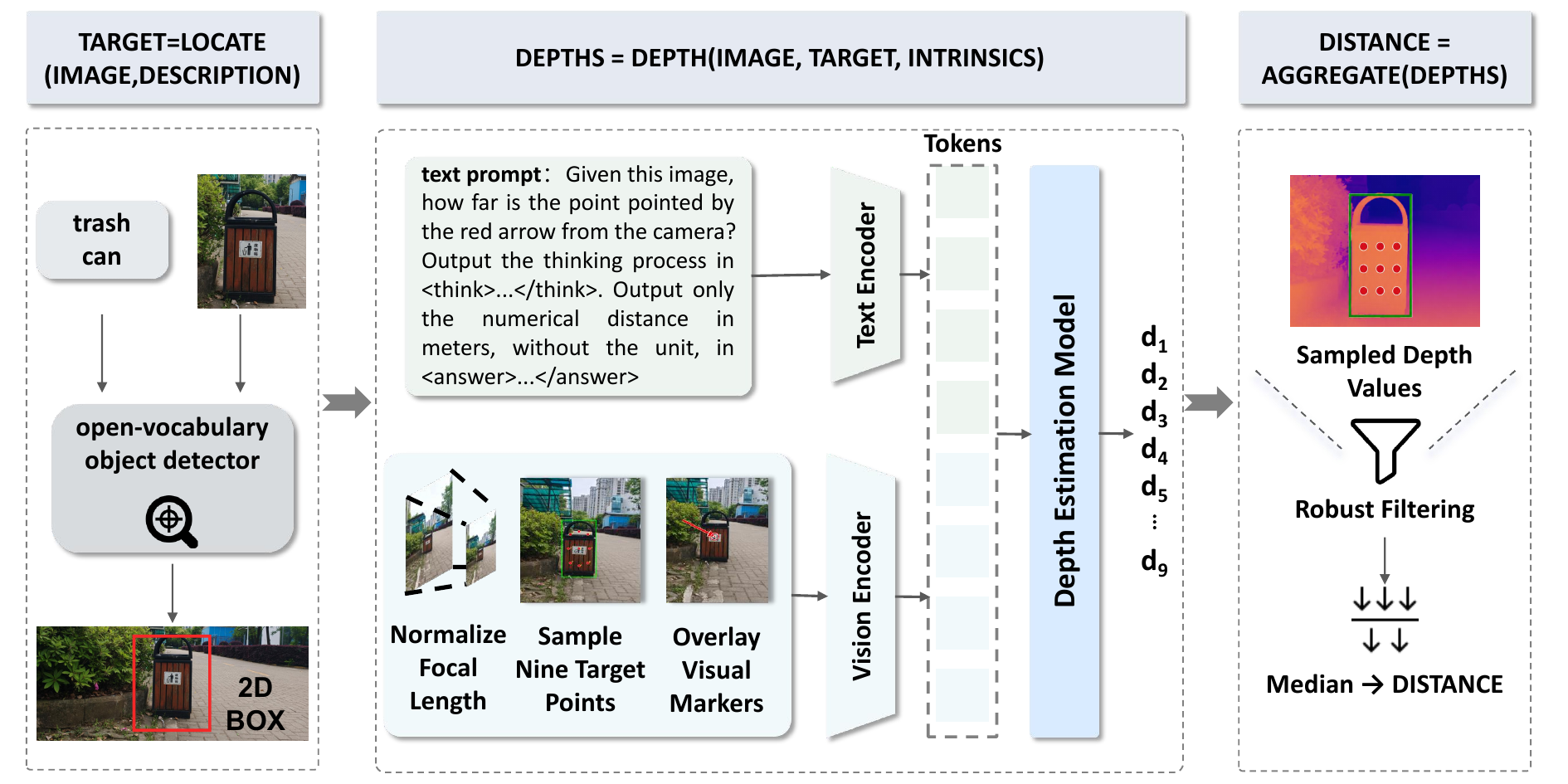}
    \caption{Overview of the instance-level metric depth estimation pipeline in M3SunAgent. \textit{LOCATE} identifies the object referred to in the input description and returns its 2D bounding box. \textit{DEPTH} normalizes the focal length and samples nine points within the box. It then uses a depth estimation model to predict the depth of these sampled points. \textit{AGGREGATE} filters unreliable predictions and computes the median of the remaining depth estimates to obtain the target's distance.
    }
    \label{fig:keypoint_sampling}
\end{figure*}

\subsection{Instance-Level Metric Depth Estimation Pipeline}
\label{sec3sub2}

For instance-level metric depth estimation, M3SunAgent generates a DSL program with three operators: \textit{LOCATE}, \textit{DEPTH}, and \textit{AGGREGATE}. As summarized in Algorithm~\ref{alg:prompted_metric_depth}, the first LLM invocation identifies the requested task and generates the program. The interpreter then executes it using registered tools. These tools localize the referred object, estimate metric depth at selected points within its image region, and aggregate the predictions into an instance-level metric distance. The second LLM invocation uses this distance and the original query to formulate the final answer. Fig.~\ref{fig:keypoint_sampling} illustrates the detailed execution process of the DSL program. The specific details are given below.

\textbf{LOCATE} The \textit{LOCATE} operator takes the input image and the target description extracted during task planning. It invokes a registered localization tool built on an open-vocabulary object detector~\cite{GroundingDINO}. Given \textit{IMAGE} and \textit{DESCRIPTION}, the tool identifies the referred instance and returns its 2D bounding box as \textit{TARGET}. This step focuses subsequent depth sampling on the target region.

\begin{figure*}[!t]
    \centering
    \includegraphics[width=0.95\linewidth]{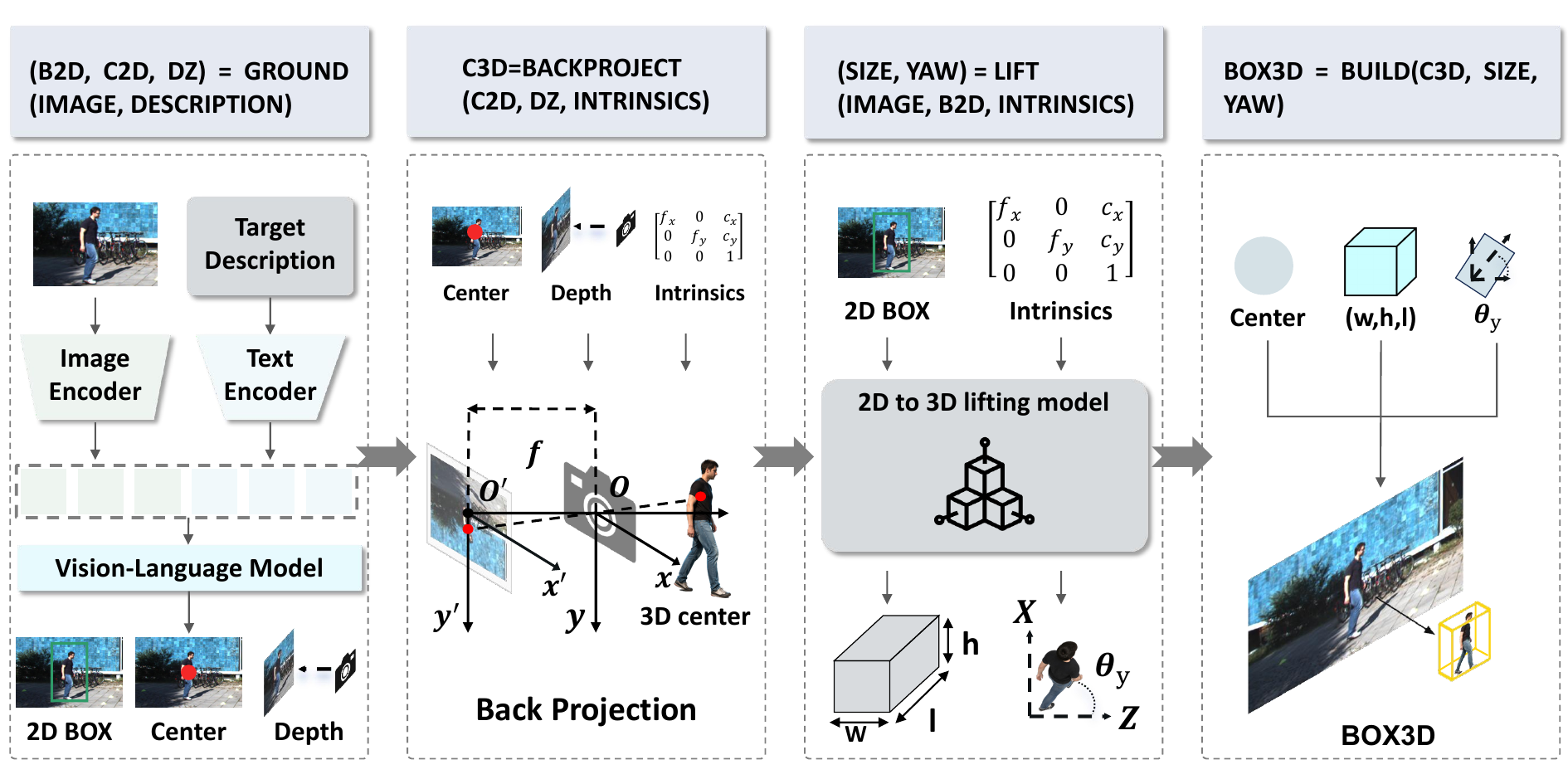}
    \caption{Overview of the monocular 3D visual grounding pipeline in M3SunAgent. Given an RGB image and a target description, \textit{GROUND} predicts the referred object's 2D bounding box, the image-plane projection of its 3D center, and its depth. \textit{BACKPROJECT} uses the projected center, depth, and camera intrinsics to recover the 3D center. \textit{LIFT} estimates the object's dimensions and yaw angle from the image, 2D box, and camera intrinsics. \textit{BUILD} combines the recovered attributes to construct the final 3D bounding box.}
    \label{fig:monocular_3d_grounding}
\end{figure*}

\textbf{DEPTH} Instance-level metric depth estimation requires the point-wise depth predictions to be associated with the referred instance. When executing \textit{DEPTH}, the interpreter invokes a registered depth estimation model~\cite{depthlm}. Given \textit{IMAGE}, \textit{TARGET}, and \textit{INTRINSICS}, the tool applies focal length normalization and estimates metric depth at multiple locations within the target region.

Metric depth prediction is sensitive to variations in camera focal length~\cite{wu2020block}. To reduce this sensitivity, the input image is resized to normalize its effective focal length to a canonical value.  The target bounding box is then used to sample nine points: one center point, four axial points, and four points near the corners. This pattern covers different parts of the target and avoids relying on a single location.

A visual marker is placed at each sampled point, and the resulting marked images are passed to the depth model. The model predicts the metric depth at each location and returns the nine values as \textit{DEPTHS}. Using a small set of representative points provides spatial coverage of the target while keeping the number of model evaluations limited.

\textbf{AGGREGATE} A 2D bounding box provides only an approximate target region and may include background pixels. Sampled points near object boundaries or occluded regions may therefore produce depths that do not represent the referred instance. The \textit{AGGREGATE} operator invokes a registered aggregation tool to reduce the influence of these unreliable depth predictions. The tool first removes non-finite values and negative depth predictions. If fewer than four valid values remain, outlier removal is skipped because quartile statistics may be unreliable for such a small sample. Let $a$ and $b$ denote the 25th and 75th percentiles of the valid depth predictions, respectively. The range of retained values is defined as
\begin{equation}
\left[
a-1.5(b-a),
b+1.5(b-a)
\right].
\label{eq}
\end{equation}
Values outside this range are treated as outliers and removed. The median of the remaining values provides a robust estimate of the target distance and is returned as \textit{DISTANCE}.

\subsection{Monocular 3D Visual Grounding Pipeline}
\label{sec3sub3}

Monocular 3D visual grounding requires identifying a language-referred object and recovering its 3D bounding box in camera coordinates. For this task, the LLM generates a DSL program comprising four operations. \textit{GROUND} predicts the target's 2D bounding box, projected 3D center, and depth. \textit{BACKPROJECT} uses camera intrinsics to recover its 3D center from these estimates. \textit{LIFT} estimates the target's dimensions and orientation. Finally, \textit{BUILD} combines these attributes into a 3D bounding box. The interpreter executes the operations using registered tools and passes intermediate results between them. Fig.~\ref{fig:monocular_3d_grounding} illustrates the detailed execution process. Algorithm~\ref{alg:monocular_3d_box} summarizes how the program is generated and executed.

\textbf{GROUND} For monocular 3D grounding, the target description may also contain spatial information, such as its relative position, distance, and spatial relationships with other objects. Such information can provide useful cues for estimating the target's 3D properties. To make better use of these semantic cues during target localization, we do not rely solely on an open-vocabulary detector as in Section~\ref{sec3sub2}. Instead, we employ a VLM~\cite{qwen3} that predicts the target's 2D bounding box while also estimating several basic spatial attributes, including the image-plane projection of its 3D center and its depth. These outputs are returned as \textit{B2D}, \textit{C2D}, and \textit{DZ}, respectively.

The bounding box \textit{B2D} is represented as $[x_1,y_1,x_2,y_2]$, and the projected center \textit{C2D} is represented as $[u,v]$. The depth \textit{DZ} denotes the depth of the target center in the camera coordinate system. The projected center is not necessarily the geometric center of the 2D bounding box. Instead, it corresponds to the projection of the target 3D center onto the image plane. According to the pinhole camera model, the target 3D center $(X_c,Y_c,Z_c)$ and its projected coordinates $(u,v)$ satisfy

\begin{equation}
u=f_x\frac{X_c}{Z_c}+c_x,
\qquad
v=f_y\frac{Y_c}{Z_c}+c_y,
\label{eq:projection_center}
\end{equation}
where $f_x$ and $f_y$ are the focal lengths in pixels along the horizontal and vertical directions, respectively, and $(c_x,c_y)$ denotes the camera principal point.

\textbf{BACKPROJECT} The \textit{BACKPROJECT} operator invokes a registered geometric tool to recover the target 3D center. Given \textit{C2D}, \textit{DZ}, and \textit{INTRINSICS}, the tool performs deterministic back-projection as

\begin{equation}
X_c=\frac{u-c_x}{f_x}d_Z,
\qquad
Y_c=\frac{v-c_y}{f_y}d_Z,
\qquad
Z_c=d_Z.
\label{eq:backprojection_center}
\end{equation}

The recovered 3D center $(X_c,Y_c,Z_c)$ is returned as \textit{C3D}. The explicit back-projection directly incorporates camera intrinsics into the computation of the 3D center. This allows the recovered coordinates to adapt to different camera configurations rather than relying on fixed camera parameters.

\textbf{LIFT} The remaining geometric attributes are estimated by the \textit{LIFT} operator. It invokes a registered lifting tool built on a specialized 2D-to-3D model~\cite{wilddet3d}. Given \textit{IMAGE}, \textit{B2D}, and \textit{INTRINSICS}, the tool estimates the object dimensions and 3D rotation. The dimensions are represented as $s=(w,h,l)$ in meters, where $w$, $h$, and $l$ denote the width, height, and length, respectively. The rotation is represented by a quaternion $(q_w,q_x,q_y,q_z)$.

In our 3D box representation, the target orientation is expressed as a yaw angle around the $Y$-axis of the camera coordinate system. Under the quaternion convention used in our implementation, the yaw angle is computed as
\begin{equation}
\theta_Y =
\operatorname{atan2}
\left(
2(q_wq_y-q_xq_z),
1-2(q_y^2+q_z^2)
\right).
\label{eq:quaternion_to_yaw}
\end{equation}

The estimated dimensions and yaw angle are returned as \textit{SIZE} and \textit{YAW}. Using the lifting tool reduces the amount of 3D geometric information that must be predicted directly by the VLM.

\textbf{BUILD} Finally, the \textit{BUILD} operator combines \textit{C3D}, \textit{SIZE}, and \textit{YAW} through a deterministic geometric operation to construct the target's 3D bounding box. The constructed box is stored as \textit{BOX3D}. In the second LLM invocation, the LLM uses this result and the original query to formulate the final answer.

\section{Experiments}
\label{sec4}

\subsection{Experimental Settings}
\label{sec4sub1}

\subsubsection{Datasets}

\begin{figure}[!htbp]
    \centering
    \includegraphics[width=\columnwidth]{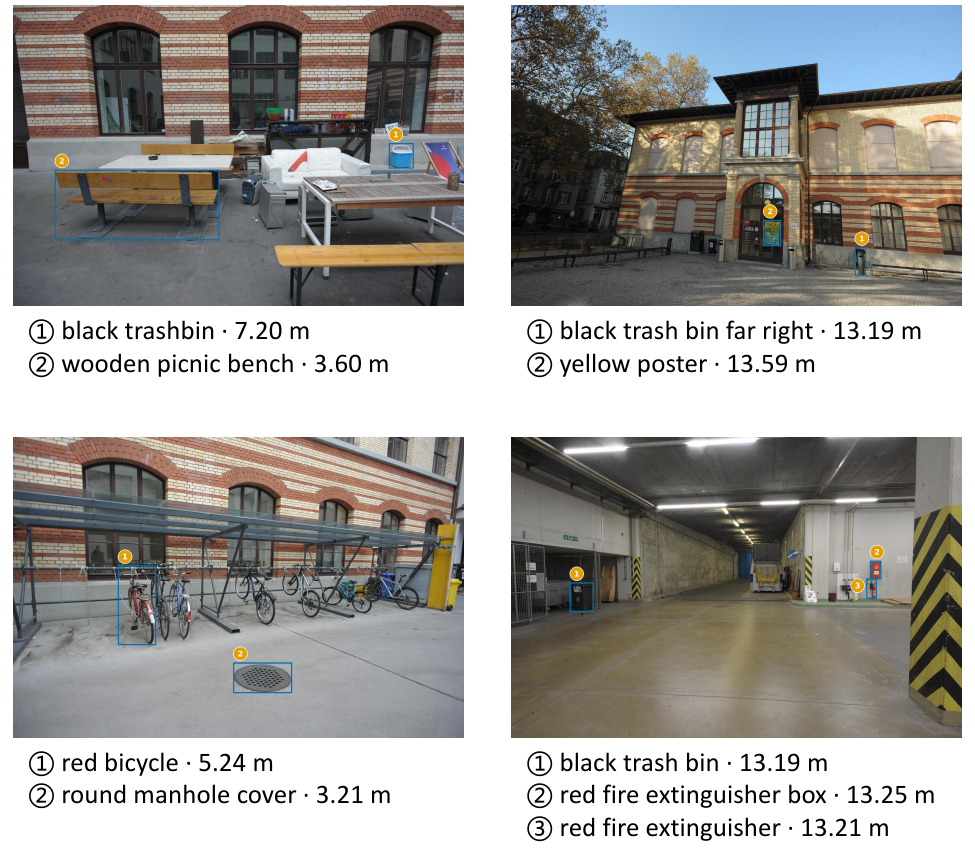}
    \caption{Examples of annotated instances in M3SI. Blue bounding boxes mark the target objects. The text beneath each image provides their descriptions and ground-truth distances from the camera in meters.}
    \label{fig:msid_examples}
\end{figure}

\indent\textbf{M3SI} To evaluate our framework on instance-level metric depth estimation, we construct a new benchmark, named the M3Sun Instance (M3SI) dataset, based on ETH3D~\cite{ETH3D}. ETH3D provides images and pixel-level depth annotations for 13 diverse real-world scenes. Although it is widely used for monocular depth estimation and stereo matching, its original annotations are not directly applicable to instance-level distance estimation, which aims to measure the metric distance from each object instance to the camera. We adapt ETH3D to this task through a multi-stage annotation pipeline. A LLM is used to identify candidate object categories and generate object descriptions for each image. When multiple objects of the same category appear in an image, attributes such as color, spatial location, and relative size are added to distinguish individual instances. The resulting object descriptions are then used as text prompts for Grounding DINO~\cite{GroundingDINO} to obtain the corresponding 2D bounding boxes. Given the detected boxes and the associated depth maps, we sample valid depth values within each detected object region. The sampled values represent axial depth along the camera $Z$-axis rather than the Euclidean distance from the camera center. We therefore use the camera intrinsics to convert each axial depth value into a Euclidean distance. The median of the sampled distances is used as the ground-truth distance for each object instance. Following this pipeline, we construct M3SI, which contains 2,910 samples for instance-level metric depth estimation. Representative annotated instances are shown in Fig.~\ref{fig:msid_examples}.

\indent\textbf{Mono3DRefer} We evaluate M3SunAgent on the \textbf{Mono3DRefer} dataset \cite{mono3dvg} to assess its performance on monocular 3D visual grounding. It contains 29,990 training samples, 5,735 validation samples, and 5,415 test samples. We use the training split to fine-tune the VLM responsible for target localization in the 3D visual grounding pipeline. We then evaluate the resulting model on the test split following the evaluation protocol defined in the original Mono3DRefer study. Following the subset definitions of Qu et al.~\cite{monovlm}, the test set is organized along three dimensions: target uniqueness, target depth, and localization difficulty. Target uniqueness divides the samples into Unique and Multiple subsets. In the Unique subset, the target is the only instance of its category in the scene, whereas the Multiple subset contains other instances of the same category. Target depth divides the samples into Near ($0$--$15$ m), Medium ($15$--$35$ m), and Far ($>35$ m) subsets. Localization difficulty follows the standard KITTI definitions~\cite{kitti}. Samples are categorized as Easy, Moderate, or Hard according to their degrees of occlusion and truncation.

\subsubsection{SOTA models for comparison}

In the metric depth estimation task, M3SunAgent is compared with two groups of models. The first group consists of general-purpose VLMs, including the closed-source model GPT-5 and the open-source models Pixtral-12B \cite{pixtral}, InternVL3-8B \cite{internvl3}, InternVL3.5-8B \cite{internvl35}, the Qwen2.5-VL series \cite{qwen25vl}, the Qwen3-VL series \cite{qwen3}, Molmo2-8B \cite{molmo2}, and LLaVA-1.5-7B \cite{llaVA}. The second group contains VLMs trained on task-specific datasets to enhance spatial perception and understanding, including SpaceLLaVA \cite{spatialvlm} and SpaceQwen \cite{SpaceQwen}, as well as DepthVLM \cite{DepthVLM}, which augments a VLM with a depth prediction head for generating dense depth maps. All models are evaluated on the same test dataset.

In monocular 3D visual grounding, we consider both VLM-based methods and computer vision (CV) methods as baselines. The VLM-based methods include MiMo-VL-7B \cite{mimovl}, Qwen2.5-VL-72B~\cite{qwen25vl}, Gemini-2.5-Flash, GPT-5, GPT-o3, Gemini-2.5-Pro, and MonoVLM \cite{monovlm}, which is specifically trained for this task. The CV methods include Mono3DVG \cite{mono3dvg}, Mono3DVG-TGE \cite{Mono3DVG-TGE} and some two-stage methods. Their results are taken from the corresponding original papers.

\begin{table*}[!htbp]
    \caption{Comparison of instance-level metric depth estimation
    results on M3SI. The best and second-best results
    are shown in bold and underlined, respectively.}
    \label{tab:metric_depth_results}
    \centering

    \footnotesize
    \renewcommand{\arraystretch}{1.05}
    \setlength{\tabcolsep}{3.2pt}

    \begin{tabular*}{0.90\textwidth}{
        @{\extracolsep{\fill}}lcccccccc@{}
    }
        \toprule

        Method
        & Year
        & \shortstack{Valid/Total}
        & MAE $\downarrow$
        & RMSE $\downarrow$
        & MedAE $\downarrow$
        & MRE (\%) $\downarrow$
        & MedRE (\%) $\downarrow$
        & $\delta_{<0.25}$ (\%) $\uparrow$
        \\

        \midrule

        \multicolumn{9}{@{}l@{}}{
            \itshape General-Purpose VLMs
        }
        \\
        \cmidrule(lr){1-9}

        GPT-5
        & 2025
        & 1970/2910
        & \underline{3.31}
        & \underline{5.07}
        & \underline{1.80}
        & \underline{39.22}
        & \underline{34.43}
        & 24.33
        \\

        Pixtral-12B~\cite{pixtral}
        & 2024
        & 2810/2910
        & 4.06
        & 6.83
        & 2.28
        & 58.25
        & 38.86
        & \underline{31.27}
        \\

        InternVL3-8B~\cite{internvl3}
        & 2025
        & 2910/2910
        & 4.60
        & 6.96
        & 2.84
        & 51.68
        & 53.06
        & 16.36
        \\

        InternVL3.5-8B~\cite{internvl35}
        & 2025
        & 2910/2910
        & 5.03
        & 7.76
        & 2.85
        & 54.24
        & 51.60
        & 20.55
        \\

        Qwen2.5-VL-3B~\cite{qwen25vl}
        & 2025
        & 2910/2910
        & 14.79
        & 82.42
        & 3.83
        & 203.50
        & 63.42
        & 17.04
        \\

        Qwen2.5-VL-7B~\cite{qwen25vl}
        & 2025
        & 2900/2910
        & 5.11
        & 10.13
        & 3.74
        & 90.34
        & 58.74
        & 22.89
        \\

        Qwen3-VL-4B~\cite{qwen3}
        & 2025
        & 2910/2910
        & 5.00
        & 7.26
        & 2.88
        & 49.85
        & 52.17
        & 16.39
        \\

        Qwen3-VL-8B~\cite{qwen3}
        & 2025
        & 2910/2910
        & 4.78
        & 6.85
        & 2.91
        & 51.54
        & 52.52
        & 14.91
        \\

        Molmo2-8B~\cite{molmo2}
        & 2026
        & 2910/2910
        & 4.84
        & 10.19
        & 2.41
        & 75.99
        & 46.64
        & 29.28
        \\

        LLaVA-1.5-7B~\cite{llaVA}
        & 2023
        & 2875/2910
        & 30.89
        & 144.69
        & 4.65
        & 463.15
        & 88.99
        & 4.91
        \\

        \midrule

        \multicolumn{9}{@{}l@{}}{
            \itshape Task-Specific Spatial VLMs
        }
        \\
        \cmidrule(lr){1-9}

        SpaceLLaVA~\cite{spatialvlm}
        & 2024
        & 2871/2910
        & 32.66
        & 53.98
        & 6.71
        & 426.52
        & 95.91
        & 2.44
        \\

        DepthVLM~\cite{DepthVLM}
        & 2026
        & 2910/2910
        & 5.05
        & 7.20
        & 3.02
        & 55.03
        & 56.68
        & 7.01
        \\

        SpaceQwen2.5-VL-3B~\cite{SpaceQwen}
        & 2025
        & 2910/2910
        & 6.29
        & 8.49
        & 4.28
        & 76.71
        & 80.78
        & 4.40
        \\

        \midrule

        \textbf{M3SunAgent (Ours)}
        & -- 
        & 2910/2910
        & \textbf{2.26}
        & \textbf{3.76}
        & \textbf{1.26}
        & \textbf{27.04}
        & \textbf{23.28}
        & \textbf{52.61}
        \\

        \bottomrule
    \end{tabular*}
\end{table*}

\subsubsection{Evaluation metrics}

We evaluate instance-level metric depth estimation using absolute and relative errors, threshold accuracy, and the distribution of relative errors. Let $\hat{d}_i$ and $d_i$ denote the predicted and ground-truth distances of sample $i$, respectively. The relative error of a valid prediction is

\begin{equation}
r_i=\frac{|\hat{d}_i-d_i|}{d_i}.
\label{eq:depth_error}
\end{equation}

Let $\mathcal{V}$ be the set of samples with valid numerical predictions, $N_v=|\mathcal{V}|$, and $N$ is the total number of test samples. We report mean absolute error (MAE), root mean squared error (RMSE), median absolute error (MedAE), mean relative error (MRE), and median relative error (MedRE) over $\mathcal{V}$. MRE and MedRE are expressed as percentages.

We define a metric, $\delta_{<0.25}$, which represents the percentage of samples with a relative error below $25\%$. It is calculated as follows:
\begin{equation}
\delta_{<0.25}
=
\frac{1}{N}
\sum_{i\in\mathcal{V}}
\mathbb{I}(r_i<0.25),
\label{eq:depth_accuracy_25}
\end{equation}
where $\mathbb{I}(\cdot)$ is the indicator function. Using $N$ as the denominator counts invalid predictions as incorrect.

To examine the error distribution, we divide the test outputs into five mutually exclusive categories. The first four contain valid predictions with relative errors in
\begin{equation}
\begin{aligned}
\mathcal{E}_1 &= [0,0.10),&
\mathcal{E}_2 &= [0.10,0.20),\\
\mathcal{E}_3 &= [0.20,0.25),&
\mathcal{E}_4 &= [0.25,0.50).
\end{aligned}
\label{eq:error_intervals}
\end{equation}
Their proportions are
\begin{equation}
p_k
=
\frac{1}{N}
\sum_{i\in\mathcal{V}}
\mathbb{I}(r_i\in\mathcal{E}_k),
\qquad k\in\{1,2,3,4\}.
\label{eq:error_bin_ratio}
\end{equation}

The fifth category includes valid predictions with relative errors of at least $50\%$ and all invalid predictions:
\begin{equation}
p_5
=
\frac{1}{N}
\left[
(N-N_v)
+
\sum_{i\in\mathcal{V}}
\mathbb{I}(r_i\geq 0.50)
\right].
\label{eq:error_invalid_ratio}
\end{equation}
This distribution provides a detailed view of prediction errors across different accuracy ranges while also accounting for large errors and invalid outputs.

For monocular 3D visual grounding, we measure the volumetric overlap between each predicted 3D bounding box and its ground-truth box. Let $\mathrm{IoU}^{3D}_i$ denote this overlap for sample $i$, and let $\mathcal{S}$ be the evaluation set with $N_{\mathcal{S}}=|\mathcal{S}|$. Mean 3D IoU (mIoU) is reported as a percentage:
\begin{equation}
\mathrm{mIoU}(\mathcal{S})
=
\frac{1}{N_{\mathcal{S}}}
\sum_{i\in\mathcal{S}}
\mathrm{IoU}^{3D}_i.
\label{eq:miou}
\end{equation}

\begin{figure*}[!t]
    \centering
    \includegraphics[width=\textwidth]{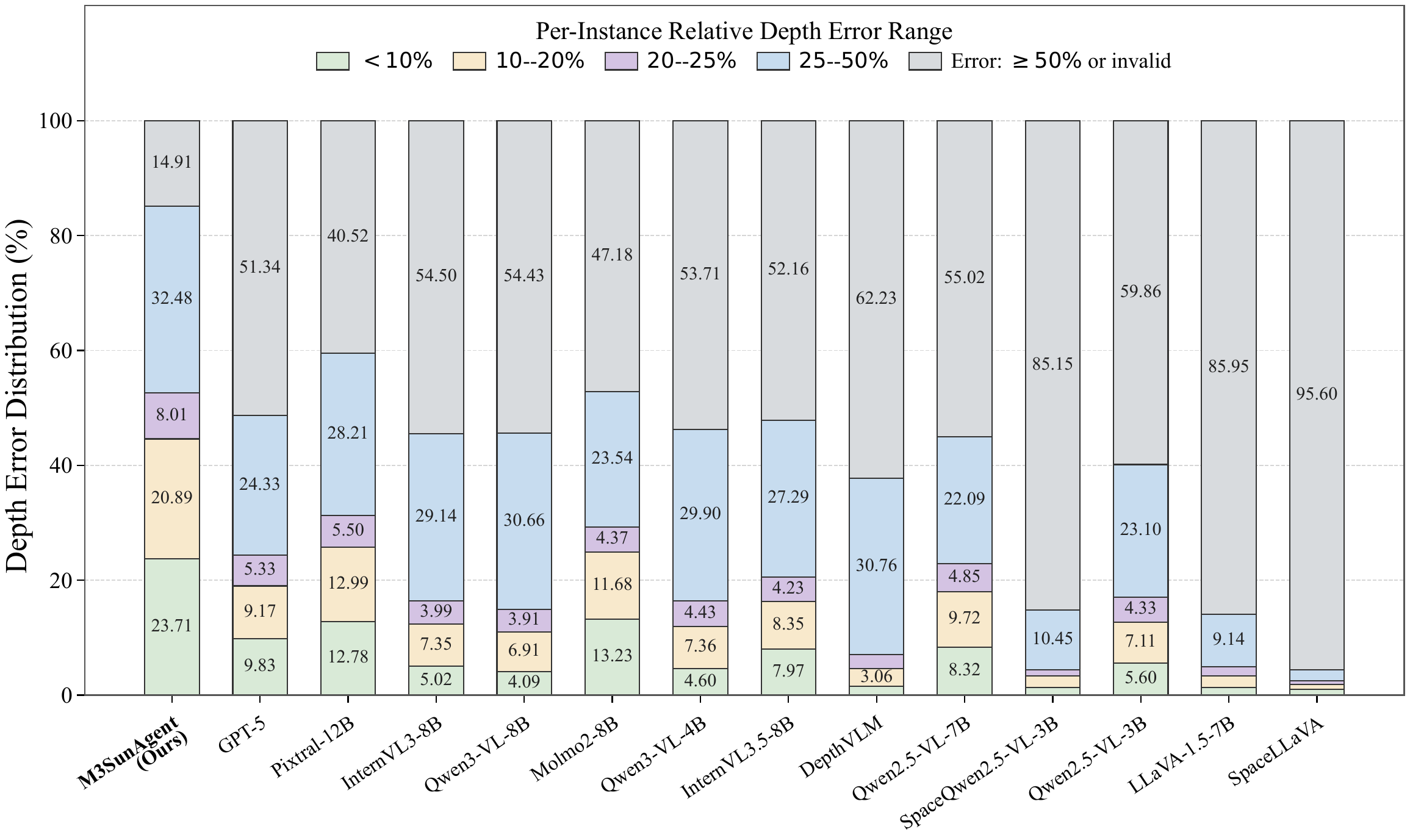}
    \caption{Distribution of per-instance relative depth errors on M3SI. Each stacked bar shows the percentage of test instances in five mutually exclusive categories: errors below $10\%$, in $[10\%,20\%)$, $[20\%,25\%)$, or $[25\%,50\%)$, and errors of at least $50\%$ or invalid predictions.}
    \label{fig:eth3d2_accuracy}


    \includegraphics[width=\textwidth]{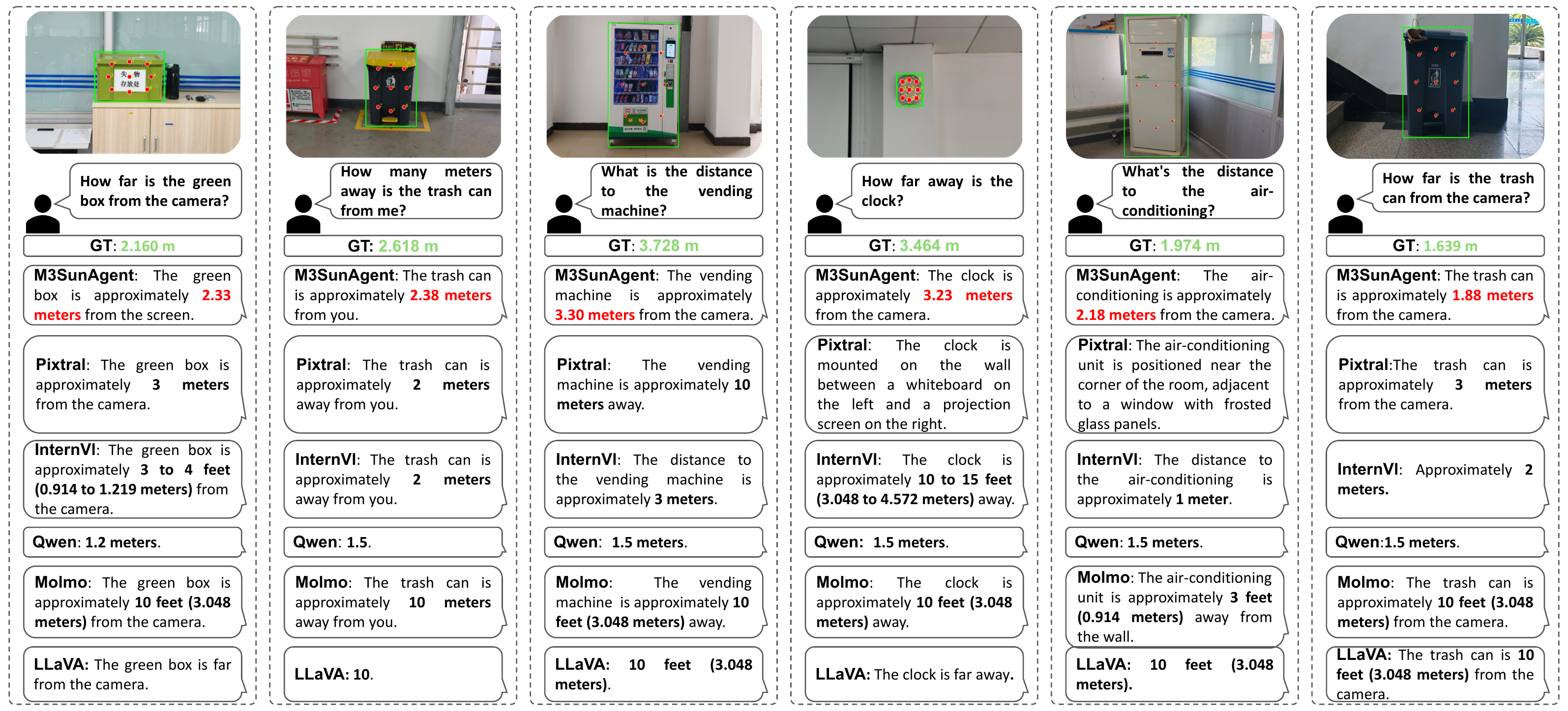}
    \caption{Qualitative comparison of M3SunAgent and general-purpose VLMs on real-world instance-level metric depth estimation. Each column shows the image and query, the ground-truth distance (GT), and the models' responses. Green boxes mark the referred objects, while red dots indicate the depth-sampling locations used by M3SunAgent.}
    \label{fig:qualitative_comparison}
\end{figure*}

A higher mIoU indicates greater overall overlap between the predicted and ground-truth 3D bounding boxes. It provides a continuous measure of 3D localization quality and reflects errors in the predicted object position, dimensions, and orientation.

We also report grounding accuracy at a given IoU threshold $\tau$:

\begin{equation}
\mathrm{Acc}@\tau(\mathcal{S})
=
\frac{1}{N_{\mathcal{S}}}
\sum_{i\in\mathcal{S}}
\mathbb{I}
\left(
\mathrm{IoU}^{3D}_i\geq\tau
\right).
\label{eq:grounding_accuracy}
\end{equation}
We report $\mathrm{Acc}@0.25$ and $\mathrm{Acc}@0.5$, which represent the percentages of samples with 3D IoU values of at least $0.25$ and $0.5$, respectively. While mIoU measures the average overlap quality, $\mathrm{Acc}@\tau$ measures the proportion of predictions that meet a specified localization threshold. The two metrics therefore provide complementary measures of 3D grounding performance.

\begin{table*}[!htbp]
    \centering
    \caption{Comparison of 3D visual grounding accuracy (\%) on the
    Mono3DRefer~\cite{mono3dvg} test set. The best and second-best results are shown
    in bold and underlined, respectively.}
    \label{tab:mono3drefer_accuracy_all}

    \footnotesize
    \renewcommand{\arraystretch}{1.05}
    \setlength{\tabcolsep}{3.2pt}

    \begin{tabular*}{0.90\textwidth}{
        @{\extracolsep{\fill}}lccccccc@{}
    }
        \toprule

        \multirow{3}{*}{Method}
        & \multirow{3}{*}{Year}
        & \multicolumn{3}{c}{$\mathrm{Acc}@0.25$}
        & \multicolumn{3}{c}{$\mathrm{Acc}@0.5$}
        \\

        \cmidrule(lr){3-5}
        \cmidrule(lr){6-8}

        &
        & Unique
        & Multiple
        & Overall
        & Unique
        & Multiple
        & Overall
        \\

        \midrule

        \multicolumn{8}{@{}l@{}}{\itshape CV Methods}
        \\
        \cmidrule(lr){1-8}

        Cube R-CNN + Rand~\cite{mono3dvg}
        & 2024
        & 32.76 & 13.36 & 17.02
        & 14.61 & 7.21 & 8.60
        \\

        Cube R-CNN + Best~\cite{mono3dvg}
        & 2024
        & 35.29 & 60.52 & 55.77
        & 16.67 & 32.99 & 29.92
        \\

        ZSGNet + backproj~\cite{mono3dvg}
        & 2024
        & 9.02 & 16.56 & 15.14
        & 0.29 & 2.23 & 1.87
        \\

        FAOA + backproj~\cite{mono3dvg}
        & 2024
        & 11.96 & 13.79 & 13.44
        & 2.06 & 2.12 & 2.11
        \\

        ReSC + backproj~\cite{mono3dvg}
        & 2024
        & 11.96 & 23.69 & 21.48
        & 0.49 & 3.94 & 3.29
        \\

        Mono3DVG~\cite{mono3dvg}
        & 2024
        & 57.65 & 65.92 & 64.36
        & 33.04 & \underline{46.85} & \underline{44.25}
        \\

        Mono3DVG-TGE~\cite{Mono3DVG-TGE}
        & 2025
        & \underline{62.45} & 69.83 & 68.44
        & \textbf{44.22} & \textbf{52.83} & \textbf{51.21}
        \\

        \midrule

        \multicolumn{8}{@{}l@{}}{\itshape General-Purpose VLMs}
        \\
        \cmidrule(lr){1-8}

        MiMo-VL-7B~\cite{mimovl}
        & 2025
        & 0.81 & 1.18 & 1.11
        & 0.00 & 0.06 & 0.05
        \\

        Qwen2.5-VL-72B~\cite{qwen25vl}
        & 2025
        & 0.69 & 0.08 & 0.20
        & 0.00 & 0.00 & 0.00
        \\

        Gemini-2.5-Flash
        & 2025
        & 0.23 & 0.86 & 0.74
        & 0.00 & 0.00 & 0.00
        \\

        GPT-5
        & 2025
        & 2.09 & 6.89 & 5.98
        & 0.36 & 0.21 & 0.23
        \\

        GPT-o3
        & 2025
        & 1.71 & 3.04 & 2.79
        & 0.19 & 0.18 & 0.18
        \\

        Gemini-2.5-Pro
        & 2025
        & 0.81 & 2.04 & 1.81
        & 0.00 & 0.09 & 0.07
        \\

        \midrule

        \multicolumn{8}{@{}l@{}}{\itshape Task-Specific VLMs}
        \\
        \cmidrule(lr){1-8}

        MonoVLM-Qwen~\cite{monovlm}
        & 2026
        & 53.99 & 63.77 & 61.89
        & 31.02 & 39.08 & 38.13
        \\

        MonoVLM-MiMo~\cite{monovlm}
        & 2026
        & 61.59
        & \underline{71.23}
        & \underline{69.41}
        & 32.68
        & 45.34
        & 42.96
        \\

        \midrule

        \textbf{M3SunAgent (Ours)}
        & --
        & \textbf{67.94}
        & \textbf{71.58}
        & \textbf{70.90}
        & \underline{36.47}
        & 44.10
        & 42.66
        \\

        \bottomrule
    \end{tabular*}
\end{table*}

\begin{table*}[!htbp]
    \centering
    \caption{Comparison of 3D visual grounding accuracy (\%) on selected
    subsets of the Mono3DRefer test set. 
    The best and second-best results are shown in bold and underlined,
    respectively.}
    \label{tab:mono3drefer_subsets}

    \scriptsize
    \renewcommand{\arraystretch}{1.00}
    \setlength{\tabcolsep}{2.2pt}

    \begin{tabular*}{0.90\textwidth}{
        @{\extracolsep{\fill}}lccccccccc@{}
    }
        \toprule

        \multirow{3}{*}{Method}
        & \multirow{3}{*}{Year}
        & \multicolumn{2}{c}{Near}
        & \multicolumn{2}{c}{Easy}
        & \multicolumn{2}{c}{Medium}
        & \multicolumn{2}{c}{Moderate}
        \\

        \cmidrule(lr){3-4}
        \cmidrule(lr){5-6}
        \cmidrule(lr){7-8}
        \cmidrule(lr){9-10}

        &
        & $\mathrm{Acc}@0.25$
        & $\mathrm{Acc}@0.5$
        & $\mathrm{Acc}@0.25$
        & $\mathrm{Acc}@0.5$
        & $\mathrm{Acc}@0.25$
        & $\mathrm{Acc}@0.5$
        & $\mathrm{Acc}@0.25$
        & $\mathrm{Acc}@0.5$
        \\

        \midrule

        \multicolumn{10}{@{}l@{}}{\itshape CV Methods}
        \\
        \cmidrule(lr){1-10}

        Cube R-CNN + Rand~\cite{mono3dvg}
        & 2024
        & 17.40 & 11.45
        & 21.12 & 11.41
        & 18.01 & 8.15
        & 17.85 & 8.01
        \\

        Cube R-CNN + Best~\cite{mono3dvg}
        & 2024
        & 67.76 & 41.45
        & 59.66 & 33.05
        & 60.69 & 30.35
        & 60.56 & 33.45
        \\

        ZSGNet + backproj~\cite{mono3dvg}
        & 2024
        & 24.87 & 0.59
        & 21.33 & 3.35
        & 16.74 & 3.71
        & 13.87 & 0.63
        \\

        FAOA + backproj~\cite{mono3dvg}
        & 2024
        & 18.03 & 0.53
        & 17.51 & 3.43
        & 15.64 & 3.95
        & 12.18 & 1.34
        \\

        ReSC + backproj~\cite{mono3dvg}
        & 2024
        & 33.68 & 0.59
        & 27.90 & 5.71
        & 24.03 & 6.15
        & 19.23 & 1.97
        \\

        Mono3DVG~\cite{mono3dvg}
        & 2024
        & 64.74 & 53.49
        & 72.36 & \underline{51.80}
        & \underline{75.44} & \underline{55.48}
        & \underline{69.23} & \underline{48.66}
        \\

        Mono3DVG-TGE~\cite{Mono3DVG-TGE}
        & 2025
        & 68.02 & \underline{56.64}
        & \underline{76.91} & \textbf{60.30}
        & \textbf{78.49} & \textbf{61.87}
        & \textbf{73.66} & \textbf{56.48}
        \\

        \midrule

        \multicolumn{10}{@{}l@{}}{\itshape General-Purpose VLMs}
        \\
        \cmidrule(lr){1-10}

        MiMo-VL-7B~\cite{mimovl}
        & 2025
        & 1.63 & 0.10
        & 0.69 & 0.00
        & 0.73 & 0.00
        & 0.99 & 0.00
        \\

        Qwen2.5-VL-72B~\cite{qwen25vl}
        & 2025
        & 0.18 & 0.00
        & 0.21 & 0.00
        & 0.24 & 0.00
        & 0.25 & 0.00
        \\

        Gemini-2.5-Flash
        & 2025
        & 1.13 & 0.00
        & 0.64 & 0.00
        & 0.40 & 0.00
        & 0.58 & 0.00
        \\

        GPT-5
        & 2025
        & 5.14 & 0.07
        & 4.81 & 0.08
        & 7.12 & 0.50
        & 7.18 & 0.38
        \\

        GPT-o3
        & 2025
        & 3.01 & 0.15
        & 2.98 & 0.18
        & 3.06 & 0.13
        & 3.33 & 0.14
        \\

        Gemini-2.5-Pro
        & 2025
        & 2.72 & 0.15
        & 1.65 & 0.12
        & 1.29 & 0.00
        & 1.24 & 0.00
        \\

        \midrule

        \multicolumn{10}{@{}l@{}}{\itshape Task-Specific VLMs}
        \\
        \cmidrule(lr){1-10}

        MonoVLM-Qwen~\cite{monovlm}
        & 2026
        & 74.06 & 49.72
        & 68.53 & 40.03
        & 54.52 & 33.08
        & 55.64 & 34.05
        \\

        MonoVLM-MiMo~\cite{monovlm}
        & 2026
        & \underline{78.15} & 52.18
        & 74.14 & 45.25
        & 63.74 & 36.89
        & 64.80 & 38.40
        \\

        \midrule

        \textbf{M3SunAgent (Ours)}
        & --
        & \textbf{84.74}
        & \textbf{64.34}
        & \textbf{77.34}
        & 50.73
        & 71.57
        & 40.00
        & 68.94
        & 43.24
        \\

        \bottomrule
    \end{tabular*}
\end{table*}

\begin{table*}[!htbp]
    \centering
    \caption{Grounding mIoU (\%) on the Mono3DRefer test set across
    object occurrence, distance, and difficulty subsets. The best
    and second-best results are shown in bold and underlined,
    respectively.}
    \label{tab:mono3drefer_miou_all}

    \footnotesize
    \renewcommand{\arraystretch}{1.00}
    \setlength{\tabcolsep}{2.0pt}

    \begin{tabular*}{0.90\textwidth}{
        @{\extracolsep{\fill}}lcccccccccc@{}
    }
        \toprule

        \multirow{4}{*}{Method}
        & \multirow{4}{*}{Year}
        & \multicolumn{9}{c}{mIoU (\%) $\uparrow$}
        \\

        \cmidrule(lr){3-11}

        &
        & \multicolumn{3}{c}{Object Occurrence}
        & \multicolumn{3}{c}{Distance}
        & \multicolumn{3}{c}{Difficulty}
        \\

        \cmidrule(lr){3-5}
        \cmidrule(lr){6-8}
        \cmidrule(lr){9-11}

        &
        & Unique
        & Multiple
        & Overall
        & Near
        & Medium
        & Far
        & Easy
        & Moderate
        & Hard
        \\

        \midrule

        \multicolumn{11}{@{}l@{}}{\itshape General-Purpose VLMs}
        \\
        \cmidrule(lr){1-11}

        MiMo-VL-7B~\cite{mimovl}
        & 2025
        & 1.45 & 1.76 & 1.71
        & 2.25 & 1.45 & 0.94
        & 0.95 & 1.47 & 2.84
        \\

        Qwen2.5-VL-72B~\cite{qwen25vl}
        & 2025
        & 1.33 & 0.79 & 0.89
        & 0.99 & 0.85 & 0.74
        & 1.00 & 0.92 & 0.72
        \\

        Gemini-2.5-Flash
        & 2025
        & 1.02 & 1.35 & 1.29
        & 1.82 & 0.94 & 0.66
        & 1.00 & 1.23 & 1.72
        \\

        GPT-5
        & 2025
        & 5.91 & 7.91 & 7.53
        & 8.16 & 6.96 & 6.94
        & 7.44 & 7.42 & 7.73
        \\

        GPT-o3
        & 2025
        & 4.83 & 6.18 & 5.93
        & 6.36 & 5.99 & 5.05
        & 5.81 & 6.29 & 5.77
        \\

        Gemini-2.5-Pro
        & 2025
        & 1.58 & 2.55 & 2.37
        & 3.25 & 1.95 & 1.14
        & 2.03 & 2.24 & 2.90
        \\

        \midrule

        \multicolumn{11}{@{}l@{}}{\itshape Task-Specific VLMs}
        \\
        \cmidrule(lr){1-11}

        MonoVLM-Qwen~\cite{monovlm}
        & 2026
        & 25.31 & 30.02 & 29.13
        & 34.35 & 25.97 & 22.58
        & 31.11 & 26.35 & 28.90
        \\

        MonoVLM-MiMo~\cite{monovlm}
        & 2026
        & \underline{33.34}
        & \underline{39.22}
        & \underline{38.11}
        & \underline{43.46}
        & \underline{34.80}
        & \textbf{31.46}
        & \underline{40.02}
        & \underline{35.69}
        & \textbf{37.68}
        \\

        \midrule

        \textbf{M3SunAgent (Ours)}
        & --
        & \textbf{38.61}
        & \textbf{42.46}
        & \textbf{41.73}
        & \textbf{54.33}
        & \textbf{40.51}
        & \underline{30.53}
        & \textbf{46.67}
        & \textbf{41.72}
        & \underline{34.83}
        \\

        \bottomrule
    \end{tabular*}
\end{table*}

\begin{figure*}[!htbp]
    \centering
    \includegraphics[width=\textwidth]{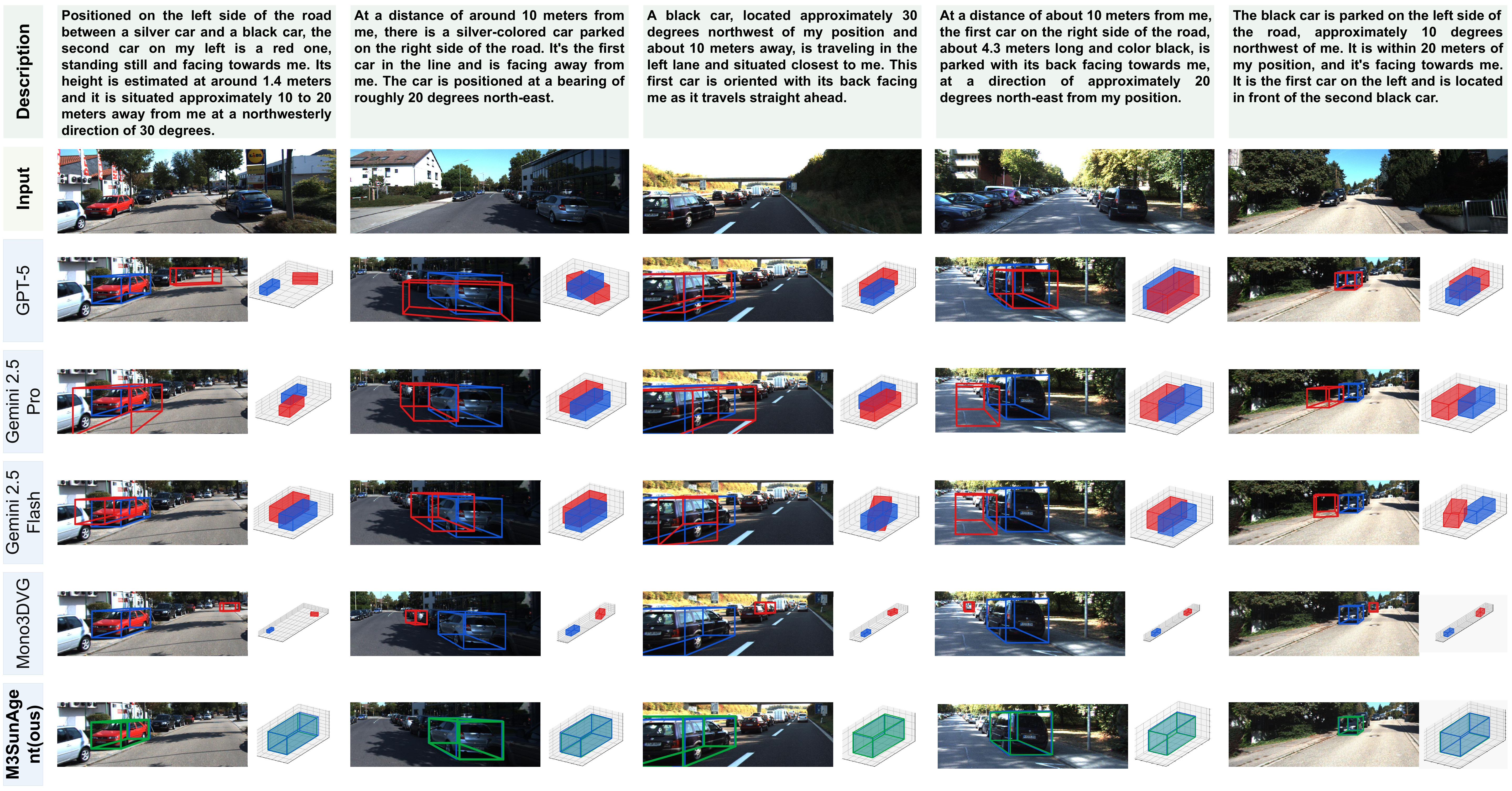}
    \caption{Qualitative comparison on five representative samples from the Mono3DRefer test set. Each column shows a referring description and an input image, followed by the 3D bounding boxes predicted by GPT-5, Gemini-2.5-Pro, Gemini-2.5-Flash, Mono3DVG, and M3SunAgent. The boxes are shown as image projections and in 3D views. Blue indicates the ground truth, red indicates baseline predictions, and green indicates M3SunAgent predictions.
    }
    \label{fig:qualitative_comparison_3D}
\end{figure*}

\subsection{Comparison with State-of-the-Art Monocular Metric Depth Estimation Methods}
\label{sec4sub2}

\subsubsection{Quantitative Evaluation} The quantitative results are reported in Table~\ref{tab:metric_depth_results} and Fig.~\ref{fig:eth3d2_accuracy}. The “Valid” column indicates the number of test samples for which a model produces a valid and parseable numerical prediction. MAE, RMSE, MedAE, MRE, and MedRE are calculated only over these valid predictions. The threshold accuracy metrics are evaluated over the entire dataset, with invalid outputs treated as incorrect predictions. The results show that many VLMs struggle to predict accurate metric depth. Some models produce large numerical errors, while others occasionally return responses that cannot be parsed. M3SunAgent achieves the best performance across all reported error metrics and produces valid predictions for all 2,910 test samples. It also obtains substantially lower errors than GPT-5. Moreover, 52.61\% of its predictions have a relative error below 25\%, demonstrating both high accuracy and stable output quality. 

Task-specific spatial VLMs do not consistently perform well on this task. SpaceLLaVA shows limited accuracy despite being trained on large-scale spatial data. DepthVLM adds a dense depth prediction head to LLM, but this design does not improve instance-level metric depth estimation on our benchmark. In contrast, M3SunAgent combines the target localization capability of an open-vocabulary detector with the metric depth estimation capability of a specialized VLM. This combination enables accurate target localization and reliable depth estimation, resulting in stable instance-level metric depth predictions across most samples. 

The stacked bar chart in Fig.~\ref{fig:eth3d2_accuracy} further compares the models under different relative error thresholds. M3SunAgent achieves the highest accuracy at all four thresholds. Under the strict 10\% threshold, it reaches 23.71\%, while the best competing model, Molmo2-8B, achieves 13.23\%. Under the more relaxed 50\% threshold, M3SunAgent achieves 85.09\%, whereas the second-best model reaches 59.48\%. These consistent gains indicate that M3SunAgent provides more accurate and reliable instance-level metric depth estimates across both strict and moderate error tolerances.

\subsubsection{Qualitative Evaluation}

Fig.~\ref{fig:qualitative_comparison} presents a qualitative comparison between M3SunAgent and general-purpose VLMs on instance-level metric depth estimation. The test images were captured with a mobile phone. The camera intrinsics were estimated from the EXIF metadata of each image. During image capture, a handheld laser rangefinder was used to measure the actual distance from the target object to the camera. This measurement was used as the ground-truth distance. The top row of the figure shows the target localization results produced by M3SunAgent. The green bounding boxes indicate the detected target regions, while the red points denote the sampling locations selected within each box. M3SunAgent can identify the referred target from the natural language query and predict a physical distance close to the laser measurement, even in real-world mobile-phone images. General-purpose VLMs can often recognize the target and understand the query. However, their distance estimates are usually coarse and rely heavily on visual and semantic priors. Some examples show large numerical errors, vague descriptions, or no valid distance value.

\subsection{Comparison with State-of-the-Art Monocular 3D Visual Grounding Methods}
\label{sec4sub3}
\subsubsection{Quantitative Evaluation}

Table~\ref{tab:mono3drefer_accuracy_all} compares M3SunAgent with the baseline methods using $\mathrm{Acc}@0.25$ and $\mathrm{Acc}@0.5$ under the Unique, Multiple, and Overall settings. Table~\ref{tab:mono3drefer_subsets} reports these metrics on the Near and Medium distance subsets and the Easy and Moderate difficulty subsets. Table~\ref{tab:mono3drefer_miou_all} presents the mIoU results across different object occurrence, target distance, and localization difficulty categories. As can be seen, M3SunAgent matches or exceeds the MonoVLM variants on most major metrics. It also remains competitive with task-specific vision methods such as Mono3DVG.

Tables~\ref{tab:mono3drefer_accuracy_all} and~\ref{tab:mono3drefer_miou_all} show that M3SunAgent performs consistently under both the Unique and Multiple settings. On the Unique subset, it achieves an $\mathrm{Acc}@0.25$ of $67.94\%$ and an mIoU of $38.61\%$. On the Multiple subset, M3SunAgent reaches an $\mathrm{Acc}@0.25$ of $71.58\%$ and an mIoU of $42.46\%$. Its mIoU is $3.24$ percentage points higher than that of MonoVLM-MiMo. Under the stricter $\mathrm{Acc}@0.5$ criterion, M3SunAgent achieves an overall accuracy of $42.66\%$, close to the $42.96\%$ obtained by MonoVLM-MiMo.

M3SunAgent performs well on nearby and less occluded targets. On the \emph{Near} and \emph{Easy} subsets, it achieves
$\mathrm{Acc}@0.25$ scores of $84.74\%$ and $77.34\%$, together with mIoU scores of $54.33\%$ and $46.67\%$, respectively. On the \emph{Medium} and \emph{Moderate} subsets, the corresponding mIoU scores are $40.51\%$ and $41.72\%$. Its $\mathrm{Acc}@0.25$ scores are $71.57\%$ and $68.94\%$, while its $\mathrm{Acc}@0.5$ scores are $40.00\%$ and $43.24\%$, respectively. On the \emph{Far} and \emph{Hard} subsets, M3SunAgent obtains mIoU scores of $30.53\%$ and $34.83\%$. These results rank second among the VLM-based methods and are $0.93$ and $2.85$ percentage points lower than those of MonoVLM-MiMo, respectively. The decrease from \emph{Near} to \emph{Far} and from \emph{Easy} to \emph{Hard} indicates that distant targets and severe occlusion or truncation remain challenging for accurate 3D localization. Nevertheless, M3SunAgent remains competitive among the VLM-based methods across all reported distance and difficulty subsets.

\subsubsection{Qualitative Evaluation} 

Fig.~\ref{fig:qualitative_comparison_3D} presents a qualitative comparison between M3SunAgent and several representative baseline methods. MonoVLM are not involved in comparison due to lack of open-source mdoel weight. General-purpose VLMs can infer the approximate target location from the language description. Their estimates of object length, width, and height are also reasonably close to the ground truth. However, their predictions of the yaw angle and 3D center contain large errors. These errors reduce the overlap between the predicted and ground-truth 3D bounding boxes. In some cases, the predicted box is completely displaced from the target. M3SunAgent accurately identifies the referred object from detailed language descriptions and produces 3D bounding boxes with greater overlap. Its estimates of the object dimensions, 3D center, and yaw angle are close to the corresponding ground-truth values.

\section{Conclusions}
\label{sec5}

This paper presents M3SunAgent, a monocular spatial understanding agent for instance-level metric depth estimation and monocular 3D visual grounding. Given an RGB image and a natural-language query, a LLM generates an executable program that coordinates specialized visual models and geometric operations. For depth estimation, the agent localizes the referred object and aggregates pointwise predictions to obtain an instance-level metric distance. For 3D visual grounding, it combines VLM-based target localization and spatial attribute prediction with camera back-projection and specialized lifting to estimate the object's 3D bounding box. This shared framework connects language-guided target identification with quantitative spatial estimation across both tasks. We also introduce the M3SI Dataset, which contains 2,910 samples for evaluating instance-level metric depth estimation. Experiments on M3SI show that M3SunAgent achieves the best overall performance across the reported depth metrics. On Mono3DRefer, it achieves overall $\mathrm{Acc}@0.25$ and $\mathrm{Acc}@0.5$ scores of $70.90\%$ and $42.66\%$, respectively. These results demonstrate the effectiveness of coordinating specialized tools to estimate the spatial properties of language-referred objects from a single RGB image. M3SunAgent may be useful to integrate physics-driven controller and data-driven vision language action to address long horizon tasks from hybrid system point of view.

\bibliographystyle{IEEEtran}
\bibliography{refs}

\vfill
\end{document}